\documentclass{article}

\PassOptionsToPackage{numbers, compress}{natbib}

\usepackage[preprint]{neurips_2022}

\usepackage[utf8]{inputenc} 
\usepackage[T1]{fontenc}    
\usepackage{hyperref}       
\usepackage{url}            
\usepackage{booktabs}       
\usepackage{amsfonts}       
\usepackage{nicefrac}       
\usepackage{microtype}      
\usepackage{xcolor}         
\usepackage{graphicx}
\usepackage{amsfonts}
\usepackage{amsmath}
\usepackage{tabularx}
\usepackage{multirow}
\usepackage{threeparttable}
\usepackage{subfigure}
\usepackage{amssymb}

\newcommand{\rv}{\textcolor{red}}

\title{An ensemble of VisNet, Transformer-M, and pretraining models for molecular property prediction in OGB Large-Scale Challenge @ NeurIPS 2022 }

\author{%
  Yusong Wang\thanks{Work done during an internship at Microsoft Research AI4Science.}\\
  Xi’an Jiaotong University; \\ Microsoft Research AI4Science\\
  \texttt{t-yusongwang@microsoft.com} \\
  \And
  Shaoning Li\footnotemark[1]\\
  Microsoft Research AI4Science \\
  \texttt{v-shaoningli@microsoft.com} \\
  \And
   Zun Wang \\
   Microsoft Research AI4Science \\
  \texttt{zunwang@microsoft.com} \\
  \And
  Xinheng He \footnotemark[1]\\
  Microsoft Research AI4Science \\
  \texttt{v-xinhenghe@microsoft.com} \\
  \And
  Bin Shao \\
  Microsoft Research AI4Science\\
  \texttt{binshao@microsoft.com} \\
  \And
    Tie-Yan Liu \\
  Microsoft Research AI4Science \\
  \texttt{tyliu@microsoft.com} \\
  \AND
  Tong Wang\thanks{Corresponding author.} \\
  Microsoft Research AI4Science \\
  \texttt{watong@microsoft.com} \\
}

\begin{document}

\maketitle

\begin{abstract}
    In the technical report, we provide our solution for OGB-LSC 2022 Graph Regression Task. 
    The target of this task is to predict the quantum chemical property, HOMO-LUMO gap for a given molecule on PCQM4Mv2 dataset.
    In the competition, we designed two kinds of models: Transformer-M-ViSNet which is an geometry-enhanced graph neural network for fully connected molecular graphs and Pretrained-3D-ViSNet which is a pretrained ViSNet by distilling geomeotric information from optimized structures. With an ensemble of 22 models, \textit{ViSNet Team} achieved the MAE of \textbf{0.0723} eV on the test-challenge set, dramatically reducing the error by 39.75\% compared with the best method in the last year competition.
\end{abstract}

\section{Introduction}
OGB-LSC is a Large-Scale Machine Learning (ML) Challenge to exploit the power of ML for graph data.
In the competition, we aim at the PCQM4Mv2 dataset, which is a quantum chemistry dataset for predicting the HOMO-LUMO energy gap of molecules calculated at Density Functional Theory (DFT) level.
Compared with PCQM4M, PCQM4Mv2 provides optimized three-dimensional structures for training but not for validation and test sets.
Therefore, how to fully utilize the 3D molecular structures and bridge the gap between the molecular topology and its 3D geometric information is the key to tackle this task. 
By elaborately analyzing the dataset, we designed two kinds models for this competition: (a) Transformer-M-ViSNet: an improved version of Transformer-M
to better extract and exploit the geometric information with powerful vector-scalar
interactive operations derived from ViSNet for fully connected molecular graphs; (b) Pretrained-3D-ViSNet: a pretrained ViSNet by distilling hidden representations of geometric information from optimized structures to generated structures.
In addition, the physicochemical insights are employed for feature engineering and model design.
The original manuscript of ViSNet\cite{wang2022visnet} is available at https://arxiv.org/pdf/2210.16518.pdf and thus we omit some details of the model design in this technical report for simplicity.
Finally, with an ensemble of 22 models, \textit{ViSNet Team} achieved the MAE of 0.0723 eV on the test-challenge set.

\section{Preliminaries}
\subsection{Runtime Geometry Calculation of ViSNet}
\label{sec:bdvisnet}
We first demonstrate how we extract the geometric information in an elegant manner.
Inspired by the message aggregation, a sufficient way to calculate the surrounding angles between the target node and its neighbors is to compute the summation of the angles as messages passed to itself.
Such method enable to reduce the computational complexity from $\mathcal{O}(\mathcal{N}^2)$ to $\mathcal{O}(\mathcal{N})$.
Its effectiveness is proved by PaiNN \cite{schutt2021equivariant} but they solely consider the angles calculation.
Meanwhile, GemNet \cite{klicpera2021gemnet} and SphereNet \cite{liu2021spherical} has testified the great influence of dihedrals for molecular modeling, but they suffer from high computational overhead due to explicitly extract the dihedrals in quadruplet atoms, which reaches $\mathcal{O}(\mathcal{N}^3)$ complexity.
To alleviate such intolerable computational overhead and maintain the utilization of dihedrals features, we propose a strategy to calculate the dihedrals in linear time complexity.
Together with the angle calculation, we name it the \textbf{Runtime Geometry Calculation} (RGC).
The details of RGC can be referred to the preprint version of ViSNet \cite{wang2022visnet}.

\subsection{Vector-Scalar Interaction for Intersecting Space}
\label{sec:ViS-IS}
We utilize the tensor product to conduct scalar and vector interaction:
\begin{equation}
    \mathbf{\Vec v} = \mathbf{s} \bigotimes v
\end{equation}
The bold symbols $\mathbf{\Vec v}$ and $\mathbf{s}$ denote the features with high dimension, i.e., $\mathbf{\Vec v} \in \mathrm{R}^{V \times F}$, $\mathbf{s} \in \mathrm{R}^F$ and $v \in \mathrm{R}^V$. Here $F$ represents the size of hidden channel and $V$ represents the dimension of the space (e.g., 3 in Cartesian coordinate or $2l+1$ in spherical space).
It is worthy noting that in this track we solely adopt $F=3$, but it could be extended to higher dimensional space by spherical harmonics \cite{batzner20223}.
The tensor product also keeps equivaraince without adding extra bias.

After extracting the geometric information, the next confronted problem is how to effectively utilize these features.
Previous works seldom consider the message transferring paths before aggregation and simply deliver all the computed messages to the target nodes.
To this end, we propose a \underline{V}ector-\underline{S}calar \underline{i}nteraction strategy for \underline{I}ntersecting \underline{S}pace, in terms of \textbf{ViS-IS} for short, as an extension of \textbf{ViS-MP} proposed in ViSNet \cite{wang2022visnet}.
The angular information is derived from the \textbf{intersecting node} $i$, which describes the spatial structure of the neighborhood $\mathcal{N}(i)$, i.e., the target node $i$ and its 1-hop neighbors.
From this view, we could treat the extracted angular features as necessary messages for target nodes.
Meanwhile, the dihedral information is derived from the \textbf{intersecting edge} $r_{ij}$, which describes the relative positions between neighborhood $\mathcal{N}(i)$ and neighborhood $\mathcal{N}(j)$.
Similarly, the edge feature of $e_{ij}$ should not only contain the distance but be complemented with extracted dihedral features.
Therefore, the process of ViS-IS can be summarized as:
\begin{equation}
    \begin{aligned}
        h_i & \leftarrow \phi(\langle \mathbf{\Vec v}_i, \mathbf{\Vec v}_i \rangle) \\
        f_{ij} & \leftarrow \phi(\langle {\rm Rej}_{\Vec r_{ij}}(\mathbf{\Vec v}_i), {\rm Rej}_{\Vec r_{ji}}(\mathbf{\Vec v}_j) \rangle)
    \end{aligned}
\end{equation}
where $h_i$ denotes the node features, $f_{ij}$ denotes the edge features and $\phi$ denotes the non-linear update function.
The rejection of vectors with high dimension can be treated as rejecting vectors from each dimension by $\Vec r_{ij}$.

\subsection{Transformer-M}
Transformer-M \cite{luo2022one} is a powerful molecular model coping with 2D \& 3D graphs.
It is able to handle different molecular modalities sharing the same backbone architecture and alternatively activates their channels.
By unified training across different views, Transformer-M preserves the essential knowledge and obtains molecular representations.
Furthermore, it considers the 2D \& 3D molecules to be fully connected and pads them with graph token.
To preserve crucial graph features, it incorporates positional encoding including degree, shortest path and distances in 3D conformers as learnable attention biases.
Its Transformer layer can be summarized as:
\begin{equation}
    A(X^{l+1}) = \operatorname{softmax}(\frac{(W_QX^l)(W_KX^l)^{\mathsf{T}}}{\sqrt{d}} + \Psi^{\operatorname{2D}} + \Psi^{\operatorname{3D\;Distance}})
\end{equation}
These two attention biases are alternatively added with different probabilities, indicating the different training mode and enhancing the data diversity.
With it powerful design and elaborate supervised signals, Transformer-M reveals its formidable capability for molecular representation learning across different data modalities.

\begin{figure}
    \centering
    \includegraphics[width=\textwidth]{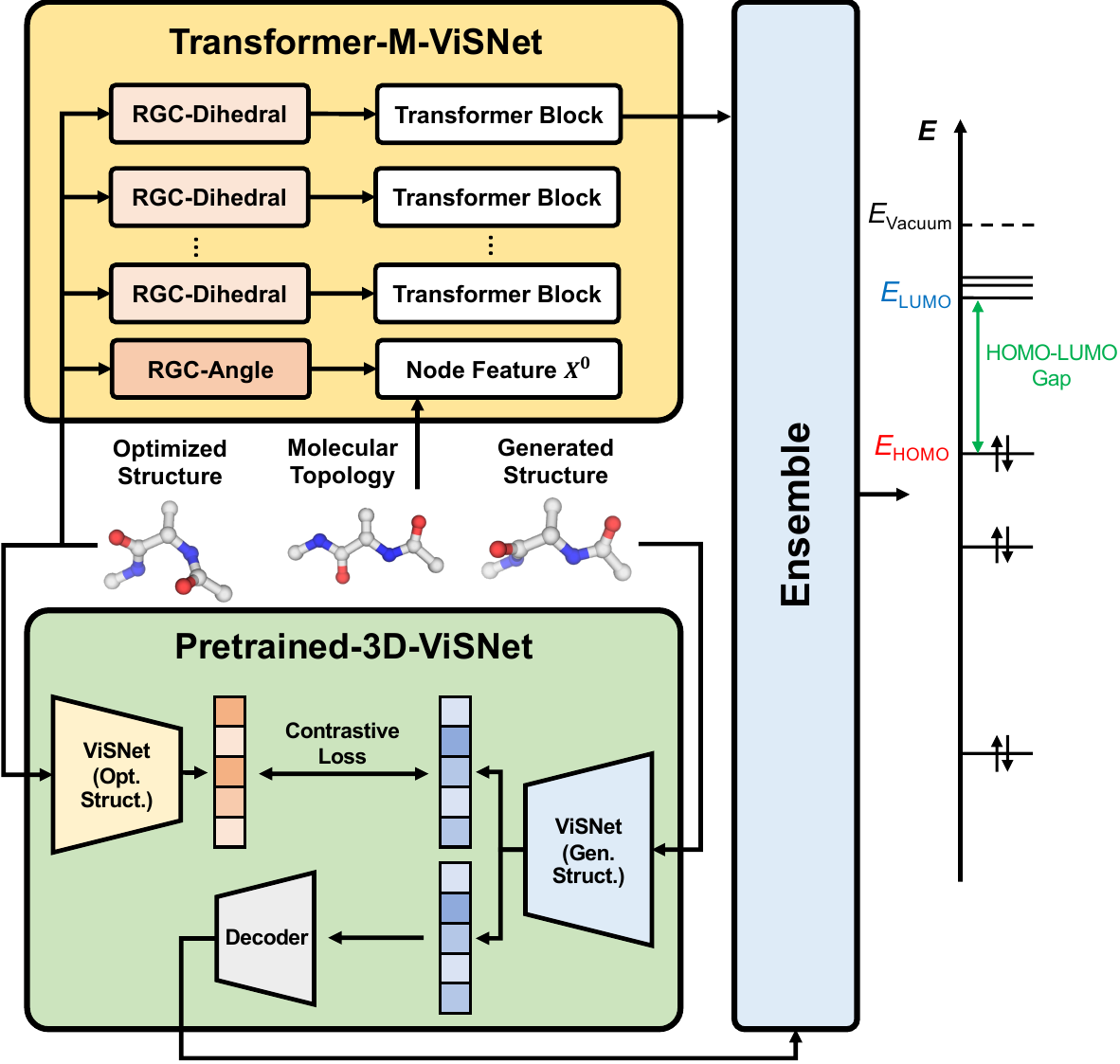}
    \caption{The overall flowchart of \textit{ViSNet Team} solution of OGB-LSC @ NeurIPS 2022 for graph regression task. Two kinds of graph models are designed for the task. The 2D molecular topology and the 3D optimized structures are fed into Transformer-M-ViSNet while both the generated and optimized structures are taken as input for Pretrained-3D-ViSNet. In Transformer-M-ViSNet, the RGC-Angle proposed by ViSNet is adopted into the node feature and RGC-Dihedral is employed as 3D attention bias for each Transformer block. In Pretrained-3D-ViSNet, the hidden representations of molecular geometric information are distilled from a ViSNet model trained with optimized structures to that trained with generated structures. The two kinds of models form an ensemble to make final HOMO-LUMO gap predictions.}
    \label{fig:arch1}
\end{figure}

\section{Methodology}
As shown in Fig.1, we propose two kinds of models for HOMO-LUMO gap prediction. The Transformer-M-ViSNet model takes the 2D molecular topology and the optimized structure as input and makes a joint training with both 2D and 3D information. The Pretrained-3D-ViSNet distills the geometric information from the optimized structure to the generated structure and decodes the hidden representations of the generated structure for prediction. Finally, an ensemble of the two kinds of models are adopted to predict the HOMO-LUMO gap. 
\subsection{Transformer-M-ViSNet}
The model architecture of Transformer-M-ViSNet is designed based on Transformer-M, which has proved to be a very powerful framework for molecule modeling.
In Transformer-M, it converts the molecules to fully connected graph and incorporates atomic distance in the attention biases to encode the graph structure.
As we demonstrated above, the geometric information including angles and dihedrals can contribute a lot to molecule modeling and promote prediction accuracy. 
To this end, we can further improve the geometry encoding process by fully leveraging the more efficient strategies proposed in ViSNet \cite{wang2022visnet}.

We first modify the input features of the node by employing the angular information taking the node as axis:
\begin{equation}
    X^{(0)} = X + \Psi^{\operatorname{2D}}  + \rv{\Psi^{\operatorname{3D\;RGC\;Angle}}}
\end{equation}
where $X$ denotes the node features, i.e., the embedding of atomic number.
$\Psi^{\operatorname{2D}}$ denotes the original extracted features from SMILES, e.g., in-degree and out-degree.
The initial vectorized features are calculated through tensor product:
\begin{equation}
    \Vec V = X \bigotimes \Vec R
\end{equation}
where $\Vec R$ represents the matrix form of $\Vec r_{ij}$ since the graph is fully connected.
Then we can further compute the additional 3D angle features using RGC:
\begin{equation}
    \rv{\Psi^{\operatorname{3D\;RGC\;Angle}}} = \langle W_{as} \Vec V, W_{at} \Vec V \rangle
\end{equation}
with $W_{as}$, $W_{at}$ as linear augmentation without bias to ensure equivariance.
We then modify the attention bias for each pair of nodes in each Transformer layer by incorporating the dihedrals taking the pair of nodes as the rotation axis. Thus the attention mechanism is defined as follows:
\begin{equation}
    A(X) = \operatorname{softmax}(\frac{XW_Q(XW_K)^{\intercal}}{\sqrt{d}} + \Psi^{\operatorname{2D}} + \Psi^{\operatorname{3D\;Distance}}+ \rv{\Psi^{\operatorname{3D\;RGC\;Dihedral}}})
\end{equation}
where $\Psi^{\operatorname{3D\;Distance}}$ denotes the sum of Euclidean distances in Transformer-M. 
The attention bias of the dihedral can be computed as:
\begin{equation}
    \rv{\Psi^{\operatorname{3D\;RGC\;Dihedral}}} = \langle W_{ds}\mathrm{Rej}_{\Vec R}(\Vec V), W_{dt}\mathrm{Rej}_{\Vec R}(\Vec V) \rangle
\end{equation}
with $W_{ds}$, $W_{dt}$ as linear augmentations without bias.
In this manner, the messages from vector-scalar interactions are transferred following ViS-IS in Section \ref{sec:ViS-IS}.
It extends the original 3D features with more geometric information at the cost of linear complexity.

\subsection{Pretrained-3D-ViSNet}
The novel model design in ViSNet enables it to fully extract the 3D geometric information for molecules and predict the corresponding quantum chemical properties. 
Considering the HOMO-LUMO gap is calculated from the optimized structure, we first evaluated the power of ViSNet for directly making predictions from the optimized structure.
For the training set of PCQM4Mv2 with 3D structures, we further split the train, validation and test sets with the ratio of 8:1:1 and trained a ViSNet with 9 building blocks and 512 hidden channels. 
Notably, in this model, only atomic types and coordinates are taken as inputs. 
Interestingly, the \textit{vanilla} ViSNet achieved the MAE of 0.0216 eV on the test set with optimized structures, which indicates that ViSNet has superior ability to extract and exploit geometric information for molecular modeling.  
However, since the 3D structures are given for the training set but are not suitable for the validation, test-dev and test-challenge sets, we propose a strategy to recover the latent space of optimized structures from the generated structures by RDKit.

\begin{table}[htbp]
\centering
\caption{Input features generated by RDKit.}
\label{table:rdkit}
\begin{tabular}{lll}
\toprule
                            & Features      & Description                         \\ \midrule
\multirow{7}{*}{Node-Level} & Atomic number & Type of atoms                       \\
                            & Aromaticity   & Where the atom is aromatic          \\
                            & Formal charge & Electrical charge                   \\
                            & Chirality tag & CW, CCW, unspecified or other       \\
                            & Degree        & Number of directly-bonded neighbors \\
                            & \# Hydrogens   & Number of bonded hydrogen atoms     \\
                            & Hybridization & S, SP2, SP3, SP3D or SP3D2          \\ \midrule
\multirow{3}{*}{Edge-Level} & Bond dir      & Begin ash, begin wedge, etc.        \\
                            & Bond type     & Single, double, triple or aromatic  \\
                            & In-ring       & Whether the bond is part of a ring  \\ \bottomrule
\end{tabular}
\end{table}
Similar to CLIP \cite{radford2021learning}, the optimized structures and generated structures by RDKit can be treated as two modalities for describing the molecular quantum properties.
However, the difference is that the generated structures have much more noises and thus affect the model to construct the molecular embedding spaces accurately.
To alleviate this problem, we regard the optimized structure as the main modality and leverage it for training the model with generated structures.
Therefore, in Pretrained-3D-ViSNet, we trained the \textit{vanilla} ViSNet first by taking the optimized structures as inputs. 
We then froze the model and employed it to train a learnable ViSNet by taking the generated structures as inputs.
We applied InfoNCE or $L_1$ loss for the corresponding graph embeddings between the two modalities.
Our motivation is to approximate the latent space of ViSNet with generated structures to that of the model with optimized structures.
The graph embedding is fed into a decoder, i.e., two layers of MLP to obtain the final prediction.
The model architecture can be referred to Sec.~\ref{sec:bdvisnet} and the input features of ViSNet with generated structures are described in Table~\ref{table:rdkit}.

\section{Training Strategy}
We trained 20 variants of Transformer-M-ViSNet with different hyper-parameters.
Both the train and validation sets were taken as training samples and the corresponding configurations and differences among the variants are shown in Table \ref{table:Transformer-M-ViSNet-cfg}.
It is worth noticing that we adopted two kinds of modes (termed as "encoder version") to initialize the vectorized features, $\Vec V$. 
\textit{Encoder V1} adopts the features of the source node ($X$) and makes tensor product with $\Vec R$ as shown in Eq.5 to initialize the vectorized features of nodes.
As a contrast, \textit{Encoder V2} concatenates the features of both the source and target nodes to form the node feature $X$ before the tensor product operation.
All 20 variants of Transformer-M-ViSNet were trained on 4 NVIDIA A100 GPUs with 80 GB memory.

\begin{table}[htbp]
\centering
\caption{Hyper-parameters and configurations of Transformer-M-ViSNet}
\label{table:Transformer-M-ViSNet-cfg}
\begin{tabular}{ll}
\toprule
                   & Transformer-M-ViSNet           \\ \midrule
\#Transformer blocks           & {[}12, 18{]}                   \\
\#Hidden dimension & 768                            \\
\#Attention heads  & 32                             \\
Lr      & {[}1e-4, 2e-4{]}               \\
Embedding dropout  & 0.0                            \\
Activation dropout & 0.1                            \\
Attention dropout  & 0.1                            \\
Drop path          & 0.1                            \\
Noise scale        & 0.2                            \\
Mode distribution  & {[}(0.2,0.2,0.6), (0.2,0.6,0.2){]} \\
Encoder Version    & {[}v1, v2{]}                   \\
\# Warmup steps     & 150000                         \\
\# Total steps      & 1500000                        \\ \bottomrule
\end{tabular}
\end{table}

For Pretrained-3D-ViSNet, we generated the 3D molecular structures by RDKit and extracted the physicochemical features as shown in Table~\ref{table:rdkit}.
Due to the limitation of time and computational resources, we trained two variants on 16 NVIDIA V100 GPUs with 32 GB memory. The two models shared the same hyperparameters as shown in Table~\ref{table:Pretrained-cfg} except for different seeds.

\begin{table}[htbp]
\centering
\caption{Hyper-parameters and configurations of Pretrained-3D-ViSNet}
\label{table:Pretrained-cfg}
\begin{tabular}{ll}
\toprule
                   & Pretrained-3D-ViSNet \\ \midrule
\#GNN blocks      & 9                    \\
\#Hidden dimension & 512                  \\
\#Attention heads  & 32                    \\
Lr                 & 1e-4                 \\
Lr factor          & 0.8                  \\
Lr min             & 1e-7                 \\
Lr patience        & 15                   \\
\# Warmup steps     & 10000                \\
\# Total epochs     & 300                  \\ \bottomrule
\end{tabular}
\end{table}

\section{Experimental results}
As shown in Table 4, we collected 22 variants from two kinds of models, Transformer-M-ViSNet and Pretrained-3D-ViSNet. 
For each sample in the test-challenge set, we made predictions by all 22 models, respectively. 
We then sorted the predicted results and picked the average of the middle 10 results for each sample.
In addition, since dihedrals are important geometric information for ViSNet, for molecules with less than 4 atoms, we generated the 3D structure by RDKit and then optimized and calculated the HUMO-LOMO gap by PySCF with B3LYP functional and 6-31G* basis. 
Since these molecules are only occupied 0.04\% of the whole test set, this process can be quickly completed within 90 minutes on a single AMD EPYC 7V13 64-core CPU.   Finally, the solution of our \textit{ViSNet Team} achieved the MAE of 0.0723 eV on the test-challenge set. 
With two different kinds of model designs that fully utilize the 3D
molecular structures, we hope our solution can facilitate to bridge the gap between the molecular topology and the 3D structures and thus benefits the research on molecular modeling.

\section*{Aknowledgement}
We acknowledge Prof. Di He at Peking University for the constructive discussions and suggestions.

\bibliographystyle{abbrv}
\bibliography{mybib}

\end{document}